%% file: main.tex
\documentclass[runningheads]{llncs}
\usepackage{hyperref}
\usepackage{graphicx}
\usepackage{comment}
\usepackage{amsmath,amssymb} 
\usepackage{color}

\usepackage{url}

\usepackage{booktabs}
\usepackage{lipsum}
\usepackage{microtype}
\usepackage{nicefrac}
\usepackage{verbatim}
\usepackage{makecell}
\usepackage{bm}
\usepackage{multirow}
\usepackage{subcaption}
\usepackage{authblk}

\begin{document}
{\onecolumn
\noindent \vspace{1cm}

\noindent \textbf{\huge{MINI-Net: Multiple Instance Ranking Network for Video Highlight Detection}}

\vspace{2cm}

\noindent {\LARGE{Fa-Ting Hong, Xuanteng Huang, Wei-Hong Li, Wei-Shi Zheng}}

\vspace{2cm}



\noindent For reference of this work, please cite:

\vspace{1cm}
\noindent Fa-Ting Hong, Xuanteng Huang, Wei-Hong Li and Wei-Shi Zheng.
MINI-Net: Multiple Instance Ranking Network for Video Highlight Detection \emph{In European Conference on Computer Vision.} 2020.  

\vspace{1cm}
\noindent Bib:\\
\noindent
@inproceedings\{hong2020mini,\\
\ \ \   title=\{MINI-Net: Multiple Instance Ranking Network for Video Highlight Detection\},\\
\ \ \  author=\{Hong, Fa-Ting and Huang, Xuanteng and Li, Wei-Hong and Zheng, Wei-Shi\},\\
\ \ \  booktitle=\{European Conference on Computer Vision\},\\
\ \ \  year=\{2020\}\\
\}
}

\pagestyle{headings}

\newcommand{\eg}{\textit{e.g.}}
\newcommand{\ie}{\textit{i.e.}}

\newcommand{\ft}{\textcolor[rgb]{0,0,0}}
\newcommand{\ftsec}{\textcolor[rgb]{0,0,0}}
\newcommand{\ftth}{\textcolor[rgb]{0,0,0}}
\newcommand{\ftfo}{\textcolor[rgb]{0,0,0}}
\newcommand{\xt}{\textcolor[rgb]{0,0,0}}
\newcommand{\xtxt}{\textcolor[rgb]{0,0,0}}
\newcommand{\wh}{\textcolor[rgb]{0,0.0,0.0}}
\newcommand{\whfi}{\textcolor[rgb]{0,0,0}}
\newcommand{\whsec}{\textcolor[rgb]{0,0,0.0}}

\newcommand{\ws}{\textcolor[rgb]{0,0,0}}
\def\ECCVSubNumber{1880}  

\title{MINI-Net: Multiple Instance Ranking Network for Video Highlight Detection} 

\titlerunning{MINI-Net}
%

\author{Fa-Ting Hong\inst{1,4,5} \and
Xuanteng Huang\inst{1} \and
Wei-Hong Li\inst{3} \and Wei-Shi Zheng\inst{1,2,5} \thanks{Corresponding author}}
\authorrunning{Fa-Ting Hong et al.}

\institute{\small School of Data and Computer Science, Sun Yat-sen University, China \and
Peng Cheng Laboratory, Shenzhen 518005, China
\and
VICO Group, University of Edinburgh, United Kingdom \and
Pazhou Lab \and
Key Laboratory of Machine Intelligence and Advanced Computing, Ministry of Education, China \\
\email{\{hongft3,huangxt57\}@mail2.sysu.edu.cn}, \email{w.h.li@ed.ac.uk}, \email{wszheng@ieee.org}
}

\maketitle

\begin{abstract}
We address the weakly supervised video highlight detection problem for learning to detect segments that are more attractive in training videos given their video event label but without expensive supervision of manually annotating highlight segments. While manually averting localizing highlight segments, weakly supervised modeling is challenging, as a video in our daily life could contain highlight segments with multiple event types, \eg, skiing and surfing. In this work, we propose casting weakly supervised video highlight detection modeling for a given specific event as a multiple instance ranking network (MINI-Net) learning. We consider each video as a bag of segments, and therefore, the proposed MINI-Net learns to enforce a higher highlight score for a positive bag that contains highlight segments of a specific event than those for negative bags that are irrelevant. In particular, we form a max-max ranking loss to acquire a reliable relative comparison between the most likely positive segment instance and the hardest negative segment instance. With this max-max ranking loss, our MINI-Net effectively leverages all segment information to acquire a more distinct video feature representation for localizing the highlight segments of a specific event in a video. The extensive experimental results on three challenging public benchmarks clearly validate the efficacy of our multiple instance ranking approach for solving the problem.


\end{abstract}

\section{Introduction}
In \xt{our} daily life, people like to share their \xt{shining moments} \wh{by posting videos} on social media platforms, such as \emph{YouTube} and \emph{Instagram}. These well-edited videos in the social media platforms can quickly attract audience and convey an owner's experience.
However, behind a well-credited video\whfi{, there} is the owner's heavy workload, as producing highlight \xt{clips} from a lengthy video \whfi{by hand} is a time-consuming and laborious task. Therefore, it would be highly demanded for developing an automated tool to cut out highlights from a lengthy video, automatically \xt{generating} a highlight short-form video. 

Recently, video highlight detection has attracted an increasing amount of attention. \whfi{E}xisting methods are mainly divided into two strategies. The first \whfi{category casts the} video highlight detection \whfi{into a} supervised learning \whfi{problem} \cite{gygli2016video2gif,yao2016highlight,jiao2018three}. Given both unedited videos and \whfi{their highlight annotations labelled manually}, a ranking net is trained to score \whfi{segments in videos such that} the highlight segments \whfi{have} higher \whfi{highlight scores} than \whfi{those} non-highlight segments in the video. For example, in \cite{gygli2016video2gif}, they \whfi{proposed} a robust deep RankNet to generate a rank list of segments according to their suitability as graphic interchange format (GIF) and \whfi{designed} an adaptive Huber loss to resist the noise effect caused by \whfi{the} outlier data. However, \whfi{these methods work} in \whfi{the} supervised \whfi{learning} manner and requires massive annotation on \whfi{highlights} in the training videos, which is hard and costly to be collected.

The second strategy treats video highlight detection as a \ftsec{weakly} supervised recognition task \cite{yang2015unsupervised,xiong2019less,wang2019unsupervised}. Given certain events' videos, they treat short-form videos as a collection of highlights, while long-form videos contain a high proportion of non-highlights. \whfi{Specially, Xiong et al. \cite{xiong2019less} designed a model that learns to predict the relations between highlight segments and non-highlight segments of the same event such that the highlight segments would have higher scores than non-highlight segments in the same event.}
\whfi{Additionally,} the work \cite{yang2015unsupervised} \whfi{employs the} auto-encoder structure to narrow the reconstruction error of segments in short-form videos, which are \whfi{considered} as highlights.
However, \whfi{video highlight detection remains as a challenging problem, as in real-world scenarios an} unedited video in social media platforms may contain highlights of more than one event, \ftth{and the above mentioned detectors that are trained on videos of target event cannot well filter out the highlights of the other events}. Without such human annotation, it is hard and indeed challenging to locate the real highlight of a target event in a video and perform specific learning. 

\begin{figure}[t]
    \begin{center}
        \includegraphics[width=0.8 \linewidth]{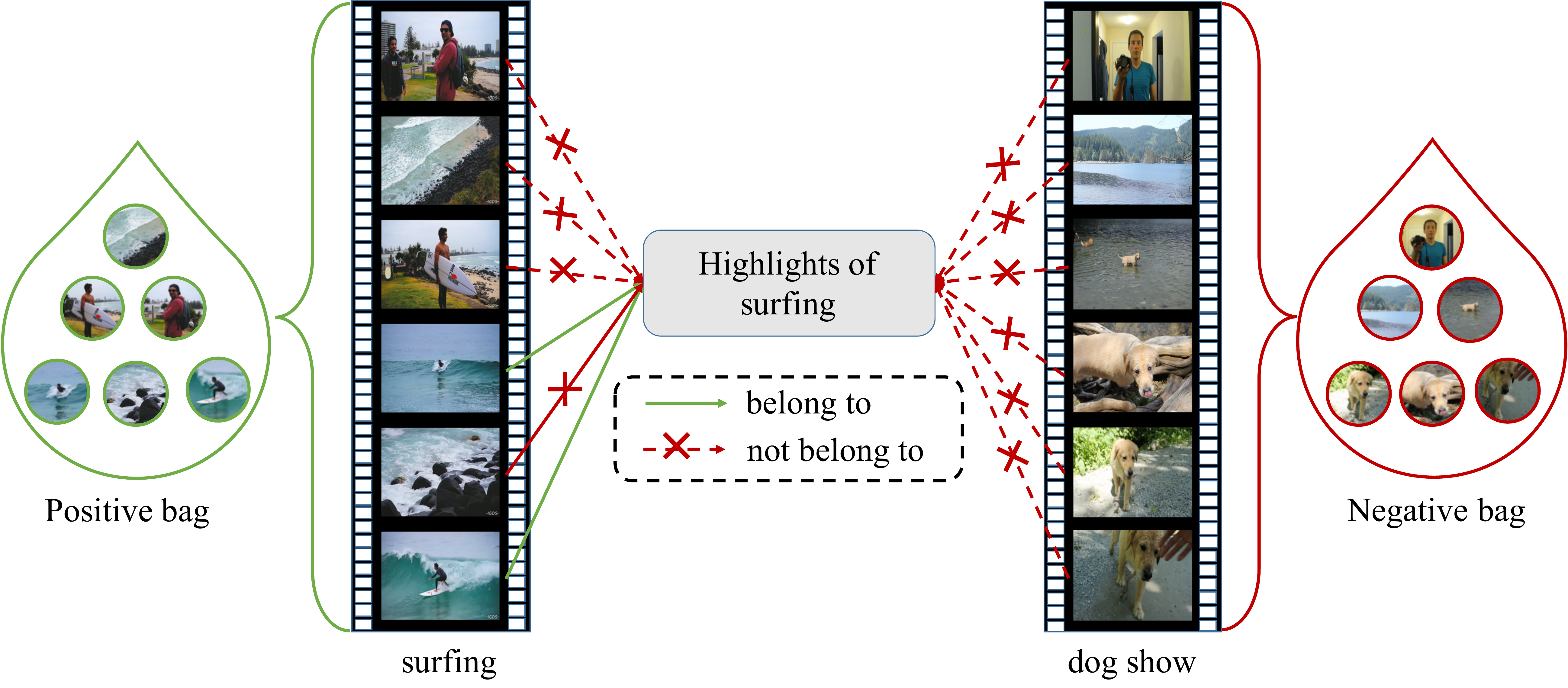}
   
      \centering\caption{\small{
      To train a model to detect surfing's highlights, we can collect unannotated videos of various events from the internet using query tags. Although the highlight annotations (\ie, labels telling which segments are highlights) are not available, we know that the videos with the ``surfing" tag (\eg, the left video) potentially contain ``surfing" highlights, while the videos of other event tags would not have highlights about surfing (\eg, videos of ``dog show" shown on the right side would not have highlights of surfing). We cast video highlight detection as a multiple instance learning problem, where we can treat videos of ``surfing" as positive bags as they contain highlights of ``surfing", while videos of other events are negative bags as they do not have highlights of ``surfing".}
    }
  \label{fig:firstpageimg}
  \end{center}
\end{figure}

\whfi{In this work, we provide a new and effective approach for solving the weakly supervised setting: even though the exact highlight annotations} of a video \whfi{are} not available, the label whether a video has a type of highlight is provided. 
In such a weakly supervised setting, we know that there exists a segment of a video that corresponds to a target highlight, but we also understand that there exist other segments of the video that do not correspond to the target highlight. To cope with this setting, we consider each video as a bag, and each bag contains a set of segments of the video (\ie, the video segments are treated as instances in each bag). Therefore, we cast the weakly supervised highlight detection as a multiple instance learning problem and proposed a Multiple InstaNce rankIng NETwork (MINI-Net) for video highlight detection.
As shown in Figure \ref{fig:firstpageimg}, for each type of highlight event, we construct positive bags using the videos \ftsec{that} that contain the target highlights \ftsec{(\eg, surfing)}, and the videos that contain other irrelevant highlight events but not the target event \ftsec{(\eg, dog show)} are used to form the negative bags. 
For such bag-level classification, we introduce two objective functions, \ie, max-max ranking loss and binary bag event classification loss, to effectively train the MINI-Net. \ws{In particular, the max-max ranking loss is designed to acquire a reliable relative comparison between the most likely positive segment instance and the most hard negative segment instance.} 
And meanwhile, minimizing the binary bag event classification loss enforces model to \ftsec{produce} more informative bag representation \ws{for the specific event}. To our best knowledge, this is the first to develop a multiple instance learning approach for weakly supervised video highlight detection.


In addition to the bag classification module, our MINI-Net also consists of two other modules: vision-audio fusion module and highlight estimation module.
The vision-audio fusion module leverages both vision features and audio features, which is beneficial as, inspired by \cite{arandjelovic2017look} learning about video segments both visually and aurally can produce more informative features.
The highlight estimation module utilizes these features to estimate the highlight score for them. We aggregate all instance features weighted by their immediate highlight scores to generate the bag feature for the bag classification module.


In our experiments, we compared the proposed model with other related methods for three challenging public video highlight detection benchmarks. \ie, YouTube Highlights dataset \cite{sun2014ranking}, TVSum dataset \cite{song2015tvsum} and CoSum dataset \cite{chu2015video}. Additionally, we have conducted an ablation study to investigate the effect of the proposed max-max ranking loss and bag classification module and validate the use of audio features and vision features. The experimental results show that our proposed model achieves a state-of-the-art performance for three public datasets and verify its efficacy for video highlight detection.

\section{Related Work}

\noindent\textbf{- Video Highlight Detection.}
In recent years, video highlight detection has attracted increasing attention. Researchers have mainly developed approaches to detect highlights of sport videos \cite{wang2004sports,xiong2005highlights,tang2011detecting} in the early stage. Recently, supervised video highlight detection has been proposed for general videos from social media platforms \cite{sun2014ranking} and first-person videos \cite{yao2016highlight}. \xt{These} methods require massive annotations for training \xt{videos which} is a time-consuming and laborious task. The Video2GIF \cite{gygli2016video2gif} method, learns from manually created GIF-video pairs, proposed a robust deep RankNet to generate a ranked list of segments according to their suitability as a GIF, and used an adaptive Huber loss to suppress the noise effect caused by outlier data. 
\ftsec{Weakly} supervised methods on video highlight detection can effectively reduce the pressure of manual labeling. More recently, methods that trained on a collection of videos of the same topic \cite{yang2015unsupervised,xiong2019less} gain a satisfactory performance. They leverage category-aware reconstruction loss \cite{yang2015unsupervised} to identify the highlights or exploit the video duration as an implicit supervision \cite{xiong2019less}.

Like \whfi{these} \ftsec{weakly} supervised video highlight detection methods, our approach also tailors highlights to the topic event. However, existing methods cannot filter the highlights of irrelevant events as they are trained on specific event videos. Unlike existing methods, our approach \xt{formulates} a multiple instance learning framework to \xt{tackle} the video highlight detection problem. Treating videos of other events as negative bags in our framework \ftsec{and using proposed max-max ranking loss to enlarge the gap between instances of target event and those of other events in terms of highlight scores} can help to filter \ftth{the} \xt{segments} of irrelevant \xt{events} and detect the highlights of the \ftth{target} event in a general video.

\noindent\textbf{- Video Summarization.} \ft{Video summarization \cite{song2015tvsum,cai2018weakly,xiong2015storyline,mahasseni2017unsupervised}, which is highly related to video highlight detection, \xt{outputs} a video summary by the estimated importance of segments. Different from video highlight detection, video summarization focuses on the integrity of the video story. Mahasseni et al. \cite{mahasseni2017unsupervised} proposed an adversarial long short-term memory (LSTM) network, consisting of a summarizer and a discriminator, to regularize the consistency between the story of the summary and the original video. In addition, by using deep reinforcement learning, \cite{zhou2018deep} formulated video summarization as a sequential decision-making process, rewarded by the diversity and representativeness of the generated video summaries. Recently, \cite{cai2018weakly} presented a generative modeling framework, which contains two important components: a variational auto-encoder for learning the latent semantics from web videos and an encoder-attention-decoder for saliency estimation of the raw video and the summary generation, to learn the latent semantic video representations to bridge the benchmark data and web data.}
Different from video summarization, our approach \whfi{selects} the highlight segments by \ftsec{comparing the instances in the training \whfi{pair}}, which \xt{consists} of \xt{one} most likely highlight an instance \xt{from} \ftth{the} positive bag and \xt{one} hard non-highlight instance \xt{from} \ftth{the} negative bag. The inherent characteristics that there is at least one positive instance in the positive bag and \xt{instances are all} negative in negative bag improve our MINI-Net's distinguishing power for detecting highlights.

\noindent\textbf{- Multiple Instance Learning.}
\ft{The multiple instance learning (MIL) is a form of weakly supervised learning in which the training instances are arranged in sets, called bags, and a label is provided for the entire bag. The field of MIL has generated a large amount of interest and is still growing \cite{wu2015deep,cinbis2016weakly,sultani2018real,ilse2018attention,carbonneau2018multiple,ulges2008multiple,meng2019weakly}. Ilse et al. \cite{ilse2018attention} proposed a neural-network-based permutation-invariant aggregation operator, a gated attention mechanism that provides insight into the contribution of each instance to the bag label, to produce bag features. Considering normal and anomalous videos as bags and video segments as instances in multiple instance learning framework, the work in \cite{sultani2018real} develops a deep multiple instance ranking framework to predict high anomaly scores for anomalous video segments.} 

\whsec{In this work, the objective of multiple instance learning is different from the above, and ours is for solving weakly supervised video highlight detection, which has not been attempted before, and some of the above MIL methods may not be applicable or effective for our problem. In addition, unlike the above MIL methods that only explore the relations among instances of a bag to encode informative bag representation and the bag classification for learning, we introduce a max-max ranking loss to acquire a reliable relative comparison between the most likely positive segment instance and the hardest negative segment instance. This enables our method for more effectively distinguishing highlight from videos, which is verified in our experiments.}

\begin{figure}[t]
    \begin{center}
        \includegraphics[width=1 \linewidth]{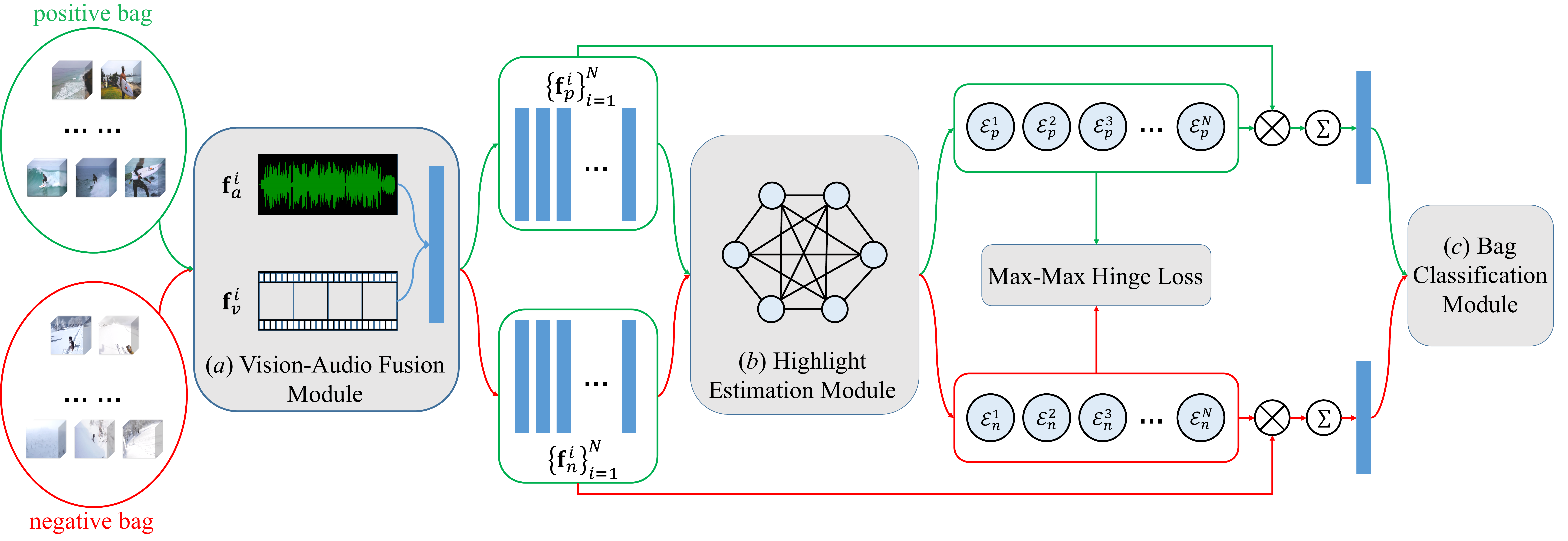}

      \centering\caption{\small{
      Illustration of our proposed MINI-Net. We feed two \xt{bags}, positive bag and negative bag, into vision-audio Fusion Module (\whfi{Figure (a)}) to \wh{encode the vision-audio fusion feature}. \wh{The highlight estimation module (\whfi{Figure (b)}) takes as input these features to estimate the highlight scores. Beyond this, the immediate highlight scores and vision-audio fusion features are fed into the bag classification module (\whfi{Figure (c)}) for bags' event category classification. The max-max ranking loss is designed to ensure that the score of the segment in the positive bag with highest score is higher than the score of the segment in the negative bag with the \xt{highest} score with a margin. Beyond this, the binary cross entropy loss is adopted for bags' event classification.}}
      }
   \label{fig:framework}
  
   \end{center}
\end{figure}
\section{Approach}

\wh{In this work, we explore event-specific\footnote{\small{We use the term event-specific to mean that there is event/category of interest specified by keyword(s) like ``surfing'', following \cite{xiong2019less,yang2015unsupervised}.}} video highlight detection under weakly supervised setting; that is we trained on unannotated data samples, in each of which the event-specific highlight exists but the annotation on its location is not specified. }
In such a weakly supervised setting, we know there exists a segment of a video corresponding to an event-specific highlight, but we also understand that there exist other segments of the video not corresponding to the event-specific highlight but probably others. Therefore, we cast the weakly supervised highlight detection as a multiple instance learning problem, and develop a Multiple InstaNce rankIng NETwork (MINI-Net) for video highlight detection. We consider each video as a bag, and each bag contains a set of segments of the video (\ie, the video segments are treated as instances in each bag). We denote the event of interest as \emph{interest event} and the other as \emph{non-interest events}, and therefore a video contains the event of interest is called a \emph{positive video} and a video that does not is called a \emph{negative video}.

More specifically, we represent a positive video as a bag $\mathcal{B}_{p}=\{\mathcal{I}_{p}^{i}\}_{i=1}^{N}$, namely a positive bag. The positive bag contains $N$ individual instances $\{\mathcal{I}_{p}^{i}\}_{i=1}^{N}$ (\ie, segments of the positive video). Similarly, the negative bag $\mathcal{B}_{n}$ contains $N$ different segments $\{\mathcal{I}_{n}^{i}\}_{i=1}^{N}$ from a negative video. \ft{Our model learns the highlights of interest event through positive bag; and through the learning of negative bag, the segments of \ftsec{the videos in non-interest events} are treated as non-highlights for the specific event.}

Given a pair of bags (\ie, a positive bag $\mathcal{B}_{p}$ and a negative bag $\mathcal{B}_{n}$), we first pre-extract the vision features $\{\mathbf{f}_v^i\}_{i=1}^N$ and audio features $\{\mathbf{f}_a^i\}_{i=1}^N$ using pretrained models. 
We then feed the pre-extracted features of both the positive bag and negative bag into the proposed model to estimate the highlight scores of instances (\ie, $\{\mathcal{E}_{p}^{i}\}_{i=1}^{N}, \{\mathcal{E}_{n}^{i}\}_{i=1}^{N}$) and event prediction (\ie, interest event or non-interest event) of two bags (\ie, $y_{\mathcal{B}_{p}}, y_{\mathcal{B}_{n}}$) as follows:

\begin{equation}\label{eq:prediction}
\begin{aligned}
	\{\mathbf{f}_{p}^{i}\}_{i=1}^{N}, \{\mathbf{f}_{n}^{i}\}_{i=1}^{N} &= f^F(\{\mathcal{I}_{p}^{i}\}_{i=1}^{N},\{\mathcal{I}_{n}^{i}\}_{i=1}^{N}|\theta^F),\\
	\{\mathcal{E}_{p}^{i}\}_{i=1}^{N},\{\mathcal{E}_{n}^{i}\}_{i=1}^{N} &= f^E(\{\mathbf{f}_{p}^{i}\}_{i=1}^{N}, \{\mathbf{f}_{n}^{i}\}_{i=1}^{N}|\theta^E),\\
	y_{\mathcal{B}_{p}}, y_{\mathcal{B}_{n}} &= f^C(\{\mathbf{f}_{p}^{i}\}_{i=1}^{N}, \{\mathbf{f}_{n}^{i}\}_{i=1}^{N}, \{\mathcal{E}_{p}^{i}\}_{i=1}^{N}, \{\mathcal{E}_{n}^{i}\}_{i=1}^{N} | \theta^C),
\end{aligned}
\end{equation}where 
$f^F(\cdot)$ is the vision-audio fusion module parameterized by $\theta^F$. The vision-audio fusion module takes each segment's vision feature and audio feature as input to encode the vision-audio fusion feature that contains both vision information and audio information (\ie, $\{\mathbf{f}_{p}^{i}\}_{i=1}^{N}, \{\mathbf{f}_{n}^{i}\}_{i=1}^{N}$ are vision-audio fusion features for the positive bag and negative bag). The encoded fusion features are input into the highlight estimation module $f^E(\cdot)$ parameterized by $\theta^E$ to predict their highlight scores. The bag classification module $f^C(\cdot)$ takes as input the vision-audio fusion features of all segments and their immediate highlight score to estimate the event category of both the positive bag and the negative bag. 

To facilitating distinguishing positive bags from negative bags, we introduce two loss functions, \ie, the max-max ranking loss and the binary bag event classification loss, to effectively train the whole multiple instance learning framework. The illustration shown in Figure \ref{fig:framework} provides an overview of our proposed method. 


\begin{figure}[t]
    \begin{center}
        \includegraphics[width=0.65 \linewidth]{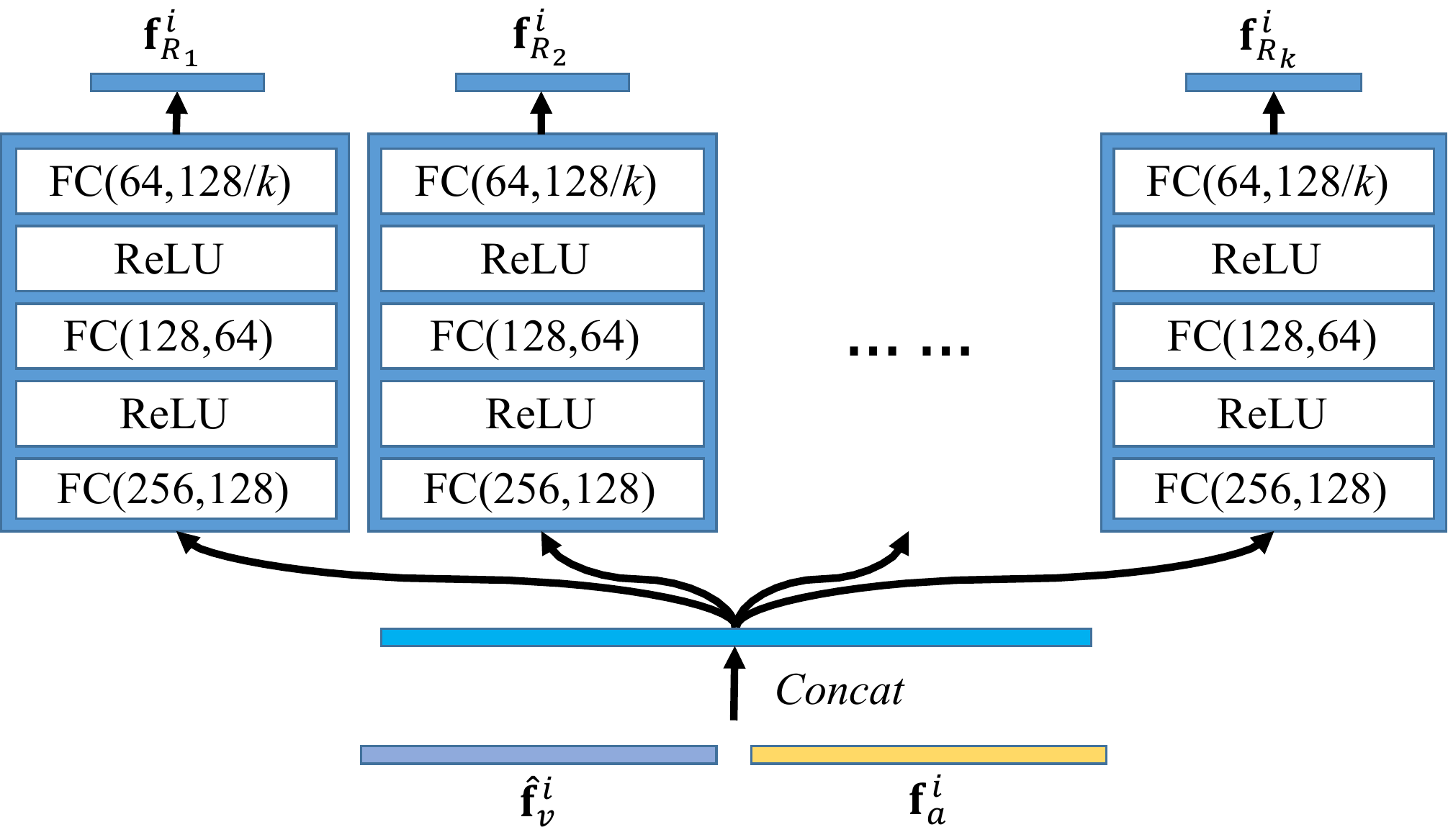}
 
      \centering\caption{\small{Illustration of the vision-audio fusion submodule. The dimension of both vision feature $\hat{\mathbf{f}}^i_v$ and audio feature $\mathbf{f}^i_a$ are 128. $k$ is the number of fusion submodule. The ``FC'' and ``ReLU'' represent fully connection and rectified linear unit activation, respectively.}
      }
   \label{fig:fusionModule}
 
   \end{center}
\end{figure}

\subsection{Vision-audio Fusion Module $f^F(\cdot)$} \label{sec:VAFM}

\wh{Given a bag of segments, instead of using the visual information to estimate the highlight score individually, we consider using both visual and audio information as visual and audio events tend to occur together, and it has been shown that audio can be adopted to assist computer vision tasks\cite{arandjelovic2017look,hori2017attention,wang2019unsupervised}.} For instance, a scene of people surfing is \wh{usually accompanied} by the sound of waves. To this end, we design a vision-audio fusion module to encode visual-audio fusion representations for video highlight detection.

\wh{Given the pre-extracted vision feature $\mathbf{f}_v^i\in\mathbb{R}^{512}$ and audio feature $\mathbf{f}_a^i\in\mathbb{R}^{128}$ of \wh{a segment} $\mathcal{I}_i$ in a bag (\ie, positive bag or negative bag), as the dimensions of both features are not the same, we first employ two fully connected layers to transform $\mathbf{f}_v^i$ to a $128$-dimensional vector, denoted as $\hat{\mathbf{f}}_v^i$. We then encode the vision-audio relation feature $\mathbf{f}_R^i$ and employ the residual connection to merge the vision-audio relation feature and vision feature $\hat{\mathbf{f}}_v^i$, yielding the vision-audio fusion feature $\mathbf{f}^i=\hat{\mathbf{f}}_v^i + \mathbf{f}_R^i$. }

To encode $\mathbf{f}_R^i$,
we concatenate the vision feature $\hat{\mathbf{f}}_v^i$ and audio feature $\mathbf{f}_a^i$ and feed the concatenated feature into $k$ parallel fusion submodules to transform the concatenated feature to $k$ relation features $\mathbf{f}_{R_k}^i$. We then concatenate the $k$ relation feature to form the vision-audio relation feature $\mathbf{f}_R^i$.

We show the architecture of the submodules in Figure \ref{fig:fusionModule}. Each fusion submodule contains 3 fully connected layers and two activation operators to transform a $256$-dimensional concatenated feature into a $\frac{128}{k}$-dimensional relation feature. In this way, the vision-audio fusion feature can be rewritten as follows:
\begin{equation}\label{eq:instanceFeat}
\begin{aligned}
  \mathbf{f}^i = \hat{\mathbf{f}}_v^i + Concat[\mathbf{f}_{R_1}^i,\dots, \mathbf{f}_{R_k}^i], (i=1,\dots,N).
\end{aligned}
\end{equation}
In this way, the $k$ parallel relation submodules allow the vision-audio fusion module to learn various types of relations between vision and audio. Additionally, encoding two sources of features (\ie, vision and audio) enables the vision-audio fusion module to automatically activate the audio information if the audio is useful for the interest event and suppress the audio information if the audio is noisy or not helpful.

\subsection{Highlight Estimation Module $f^E(\cdot)$} \label{sec:HEM}

\wh{To predict the highlight score, we feed the vision-audio fused feature $\mathbf{f}^i$ into the highlight estimation module, where we transform $\mathbf{f}^i$ into a score value that will be used \whfi{for bag classification and computing} the proposed max-max ranking loss in later sections.}
\wh{More specifically, we first compute the initial highlight score by:
\begin{equation}\label{eq:highlightvalue}
\begin{aligned}
  \hat{\mathcal{E}}^{i} = W_{H}(ReLU(W_S\mathbf{f}^{i})),
\end{aligned}
\end{equation}
where $W_S$ is a matrix projecting the vision-audio fusion feature into a subspace, the ReLU activation operator activates the effective elements, and the matrix $W_H$ is applied to measure the highlight score.}

\wh{Rather than simply using $\hat{\mathcal{E}}^{i}$ as the highlight score, we consider estimating the final highlight score using the scores of all segments in a bag since the highlight score of one segment is related to other segments in the same video. Therefore, we formulate the final score as:
\begin{equation}\label{eq:highlightScore}
  \mathcal{E}^{i} = \left(\sum^{N}_{t=1}exp(\hat{\mathcal{E}}^{t})\right)^{-1}exp(\hat{\mathcal{E}}^{i}),
\end{equation}
In this way, $\mathcal{E}^{i}$ is normalized in a bag and can be compared with the score of a segment in another bag.}

\subsection{Bag Classification Module $f^C(\cdot)$} \label{sec:BCM}

\whfi{Apart from estimating highlight scores of individual segments, we find that the event category can also be used as a supervision signal for training. \ws{The event category label can be more} easily collected as all videos \ws{can be} collected by specific query tags, and the tags can be used to generate the binary event label (\ie,  interest event or non-interest events). In addition, \ws{it} is the fact that a video may contain highlights of various events while we are only interested in a specific event's highlights. This means that correctly classifying the event category (interest event or non-interest events) can be a useful inductive bias for event-specific highlight detection.}

\whfi{\ws{More specifically}, we first label positive videos (videos of interest event) as $1$ and negative videos (videos of non-interest events) as $0$, \ie, $Y_{\mathcal{B}_p}=1$ for $\mathcal{B}_p$ and $Y_{\mathcal{B}_n}=0$ for $\mathcal{B}_n$. To classify the event category of each bag, we aggregate the 
vision-audio fusion features of all instances weighted by their immediate estimated highlight scores to generate the bag representation:
\begin{equation}\label{eq:bagfeature}
\begin{aligned}
    \mathbf{f}_{\mathcal{B}} = \sum^N_{i=1}\mathcal{E}^{i}\mathbf{f}^{i}.
\end{aligned}
\end{equation}
In this way, the generated bag representation could be highly informative for 
the event classification of each bag, as it mainly relies on the vision-audio fusion feature of the instance with high highlight scores.}

We then feed the generated bag feature $\mathbf{f}_{\mathcal{B}}$ into an event classifier that consists of two fully connected layers. We apply the softmax function to estimate the event categories for both the positive bag $y_{\mathcal{B}_p}$ and the negative bag $y_{\mathcal{B}_n}$.


\subsection{Objective Functions} \label{sec:losses}

After obtaining the predicted highlight scores of segments and the estimated event categories of the positive bag and negative bag, we introduce two objective functions (\ie, max-max ranking loss and bag event classification loss) to effectively train our MINI-Net.

\noindent\textbf{- ``max-max'' ranking loss (MM-RL).}
\whfi{To learn the highlight detection model, we expect that the highlight score of a ground-truth highlight segment is higher than the score of a non-highlight segment:
\begin{equation}\label{eq:rankingRelation}
\begin{aligned}
  \mathcal{E}_{gt-H} > \mathcal{E}_{gt-N},
\end{aligned}
\end{equation}}where $\mathcal{E}_{gt-H}$ is the highlight score of a ground-truth highlight segment and $\mathcal{E}_{gt-N}$ is the score of a non-highlight segment.

However, the highlight annotations are not available during training. 
\ws{Considering that the positive video contains at least one highlight segment, and the negative video does not have any highlights of the interest event,
}
\ws{we thus believe the segment from a positive video with the highest score is the most likely to be a highlight, and the segment from the negative bag with the highest score can be assigned as a hardest non-highlight}.
We adapt Eq. \ref{eq:rankingRelation} \ws{as follows for acquiring a reliable 
relative comparison between mostly likely positive instance and hardest negative instance}:
\begin{equation}\label{eq:adaptiveRanking}
\begin{aligned}
    \max_{\mathcal{I}_{p}^{i} \in \mathcal{B}_p} \mathcal{E}^i_p > \max_{\mathcal{I}_{n}^{i} \in \mathcal{B}_n} \mathcal{E}^i_n,
\end{aligned}
\end{equation}
where $max$ operators pick the maximum value from the highlight scores of all segments in a bag. \whfi{Here,} the highlights in \whfi{the non-interest events' videos are viewed} as non-highlights \whfi{for} the interest event. Using the segments of the non-interest events as a negative instance is more reliable than \whfi{using the segment from the long-form interest event's video \cite{xiong2019less}.}

\whfi{To instantiate Eq. \ref{eq:adaptiveRanking}, we introduce the \whfi{max-max ranking loss (MM-RL)} as:
\begin{equation}\label{eq:adaptiveLoss}
\begin{aligned}
    \mathcal{L}_{MM}(\mathcal{B}_p,\mathcal{B}_n) = \max(0,\epsilon-\max_{\mathcal{I}_{p}^{i} \in \mathcal{B}_p} \mathcal{E}^i_p + \max_{\mathcal{I}_{n}^{i} \in \mathcal{B}_n} \mathcal{E}^i_n),
\end{aligned}
\end{equation}
where $\mathcal{L}_{AH}$ is applied to ensure that $\max_{\mathcal{I}_{p}^{i} \in \mathcal{B}_p} \mathcal{E}^i_p$ is larger than $\max_{\mathcal{I}_{n}^{i} \in \mathcal{B}_n} \mathcal{E}^i_n$ with a margin of $\epsilon$. $\epsilon$ is a hyperparameter and is equal to 1 in this work.}

\noindent\textbf{- Bag \ftsec{Event} Classification Loss.} \whfi{As mentioned in Sec. \ref{sec:BCM}, in addition to the MM-RL loss, we expect the bag event classification loss can enforce the model to produce more informative bag representation \ws{for the specific event}. To this end, we apply the binary cross entropy loss function to the estimated event categories of both positive bag and negative bag for bag event classification. Finally, we add up both the MM-RL and the \ftth{bag event} classification loss to form the final loss:
\begin{equation}\label{eq:finalloss}
\begin{aligned}
    \mathcal{L} = \mathcal{L}_{MM}(\mathcal{B}_p,\mathcal{B}_n)+\mathcal{L}_{CE}(y_{\mathcal{B}_p},Y_{\mathcal{B}_p})+\mathcal{L}_{CE}(y_{\mathcal{B}_n},Y_{\mathcal{B}_n}),
\end{aligned}
\end{equation}
where $\mathcal{L}_{CE}(\cdot)$ is the binary cross entropy loss function.}


\section{Experiments}

In this section, we conduct extensive experiments on three public datasets to 
investigate the effectiveness of the proposed model. \ftsec{More experimental results and details are reported and analyzed in the Supplementary Material.}

\subsection{Datasets and Metrics} 

We evaluate our method on three public benchmarks \xt{datasets}, \ie, YouTube Highlights \cite{sun2014ranking}, TVSum \cite{song2015tvsum} and CoSum \cite{chu2015video}, for video highlight detection.

\noindent\textbf{- YouTube Highlights} contains six evnet-specific categories, \ie, dogs, gymnastics, parkour, skating, skiing and surfing, and there are approximately 100 videos in each event. The given label for YouTube highlights indicates whether a segment is a ground-truth highlight segment.

\noindent\textbf{- TVSum} is an available video summarization benchmark dataset 
that is collected from \emph{YouTube} and crawled by an event-specific queried tag. The dataset consists of 50 videos grouped by 10 categories (5 videos per category). We follow \cite{cai2018weakly,xiong2019less} and select the top $50\%$ shots in terms of the score provided by annotators for each video as a human-created summary.

\noindent\textbf{- CoSum} has 51 videos covering 10 events. We follow \cite{panda2017weakly,cai2018weakly} and compare each generated highlights with three human-created summaries.

\subsection{Compared Methods}

To further demonstrate the capacity of our method, we compare our method with numerous different methods on three datasets for video highlight detection. 

\noindent\textbf{- Weakly supervised methods}. The compared methods include RRAE \cite{yang2015unsupervised}, MBF \cite{chu2015video}, SMRS \cite{elhamifar2012see}, Quasi \cite{kim2014joint}, CVS \cite{panda2017collaborative}, SG \cite{mahasseni2017unsupervised}, and LIM-s \cite{xiong2019less}, and two weakly supervised methods, VESD \cite{cai2018weakly} and DSN \cite{panda2017weakly}. 
Although most of these methods are used for video summarization, their performance is evaluated using 
the same metrics as the metrics used in this study.

\noindent\textbf{- Supervised methods} Additionally, there are several supervised methods (\ie, GIFs \cite{gygli2016video2gif},LSVM
\cite{sun2014ranking}, KVS \cite{potapov2014category}, DPP \cite{gong2014diverse}, sLstm \cite{zhang2016video} and SM \cite{gygli2015video}) that are applied in video highlight detection and video summarization. We compare these methods using the same matrices mentioned above.


\subsection{Highlight Detection Results}\label{sec:hdResults}

\noindent\textbf{- Result for the YouTube Highlights dataset}: We report our results in comparison with other researches \footnote{\small{The compared results are from original papers.}}. \ftsec{For the sake of fairness, we also reported the results of a MINI-Net's variant, \ie, $\textup{MINI-Net}^\textup{{w/o audio}}$, which removes the audio feature from the MINI-Net and replace the vision-audio fusion feature with vision feature (more analysis about $\textup{MINI-Net}^\textup{{w/o audio}}$ is reported in Section \ref{sec:ablation}).} We find that our method achieves the best result in terms of the average mAP over all events. Compared to the ranking-based weakly supervised method LIM-s and auto-encoder-based weakly supervised method RRAE, MINI-Net's average gains in mAP are $7.96\%$ and $26.06\%$, respectively. The result strongly verifies that our weakly supervised method based on multiple instance learning has better capacity than the compared \xt{methods}. It is noteworthy that the our result is even better than that achieved by supervised methods, \ie, GIFs and LSVM, which are trained with event-specific manually annotated data. These results indicate that our MINI-Net can leverage unlabeled videos for video highlight detection more effectively than other methods without the need to spend a lot of manual labor on data annotation. \ftsec{We also find that our $\textup{MINI-Net}^\textup{{w/o audio}}$ outperforms all compared methods without audio feature. Such results indicate that proposed objective \xt{functions} can improve the ability to distinguish of our model.} 
 \begin{table}[t] 
	\centering
	   \resizebox{0.9\columnwidth}{!}
		 {
	   \renewcommand\arraystretch{0.9}
	   \begin{tabular}{l|c|c|c|c|c|c}   
		   \toprule
		   \multirow{2}{*}{Topic} &\multicolumn{2}{c|}{Supervised Methods}&\multicolumn{2}{c|}{Weakly supervised Methods} &\multicolumn{2}{c}{Weakly supervised} \\
			 \cline{2-7}
		   & GIFs  & LSVM & RRAE    & LIM-s  &  $\textup{MINI-Net}^\textup{{w/o audio}}$& MINI-Net \\
		  \hline
			dog			&$0.308$&$\mathbf{0.60}$&$0.49$&$0.579$&$0.5368$&$0.5816$ \\
			gymnastics	&$0.335$&$0.41$&$0.35$&$0.417$&$0.5281$&$\mathbf{0.6165}$\\
			
			parkour		&$0.540$&$0.61$&$0.50$&$0.670$&$0.6888$&$\mathbf{0.7020}$\\
			
			skating		&$0.554$&$0.62$&$0.25$&$0.578$&$0.7094$&$\mathbf{0.7217}$\\
			
			skiing		&$0.328$&$0.36$&$0.22$&$0.486$&$0.5834$&$\mathbf{0.5866}$\\
			
			surfing		&$0.541$&$0.61$&$0.49$&$0.651$&$0.6383$&$\mathbf{0.6514}$\\
			    
			\hline
			Average		&$0.464$&$0.536$&$0.383$&$0.564$&$0.6138$&$\mathbf{0.6436}$\\
		  \bottomrule
	   \end{tabular}
		 }%
	  
	   \caption{ \small{Experimental results (mAP) \whfi{on} the YouTube Highlights dataset. Our method outperforms all of the compared methods, including the state-of-the-art weakly supervised ranking-based method \cite{xiong2019less}.}}
        \label{tab:YouTube}

\end{table}%

\begin{table}[t] 
\centering
{
   \resizebox{0.9\columnwidth}{!}
	 {
   \renewcommand\arraystretch{0.9}
   \begin{tabular}{l|c|c|c|c|c|c|c|c|c|c|c|c|c|c}
		 \toprule
		 \multirow{2}{*}{Topic} &\multicolumn{5}{c|}{Supervised Methods}&\multicolumn{7}{c|}{Weakly supervised/Un Methods} &\multicolumn{2}{c}{Weakly supervised} \\
		 \cline{2-15}
		 &KVS & DPP & sLstm & SM &SMRS & Quasi  & MBF& CVS & SG & LIM-s& DSN &VESD& $\textup{MINI-Net}^\textup{{w/o audio}}$& MINI-Net  \\
		 \hline

		 VT &$0.353$&$0.399$&$0.411$&$0.415$&$0.272$&$0.336$&$0.295$&$0.328$&$0.423$&$0.559$ &$0.373$&$0.447$&$0.8028$&$\mathbf{0.8062}$\\
		 VU &$0.441$&$0.453$&$0.462$&$0.467$&$0.324$&$0.369$&$0.357$&$0.413$&$0.472$&$0.429$ &$0.441$&$0.493$&$0.6527$&$\mathbf{0.6832}$\\
		 GA &$0.402$&$0.457$&$0.463$&$0.469$&$0.331$&$0.342$&$0.325$&$0.379$&$0.475$&$0.612$ &$0.428$&$0.496$&$0.7535$&$\mathbf{0.7821}$\\
		 MS &$0.417$&$0.462$&$0.477$&$0.478$&$0.362$&$0.375$&$0.412$&$0.398$&$0.489$&$0.540$ &$0.436$&$0.503$&$0.8128$&$\mathbf{0.8183}$\\
		 PK &$0.382$&$0.437$&$0.448$&$0.445$&$0.289$&$0.324$&$0.318$&$0.354$&$0.456$&$0.604$ &$0.411$&$0.478$&$0.7801$&$\mathbf{0.7807}$\\
		 PR &$0.403$&$0.446$&$0.461$&$0.458$&$0.276$&$0.301$&$0.334$&$0.381$&$0.473$&$0.475$ &$0.417$&$0.485$&$0.5446$&$\mathbf{0.6584}$\\
		 FM &$0.397$&$0.442$&$0.452$&$0.451$&$0.302$&$0.318$&$0.365$&$0.365$&$0.464$&$0.432$ &$0.412$&$0.487$&$0.5586$&$\mathbf{0.5780}$\\
		 BK &$0.342$&$0.395$&$0.406$&$0.407$&$0.297$&$0.295$&$0.313$&$0.326$&$0.417$&$0.663$ &$0.368$&$0.441$&$0.7174$&$\mathbf{0.7502}$\\
		 BT &$0.419$&$0.464$&$0.471$&$0.473$&$0.314$&$0.327$&$0.365$&$0.402$&$0.483$&$0.691$ &$0.435$&$0.492$&$0.7686$&$\mathbf{0.8019}$\\
		 DS &$0.394$&$0.449$&$0.455$&$0.453$&$0.295$&$0.309$&$0.357$&$0.378$&$0.466$&$0.626$&$0.416$&$0.488$&$0.5911$&$\mathbf{0.6551}$\\
		 \hline
		 Average 	&$0.398$&$0.447$&$0.451$&$0.461$&$0.306$&$0.329$&$0.345$&$0.372$&$0.462$&$0.563$ &$0.424$&$0.481$&$0.6979$&$\mathbf{0.7324}$\\
		 \bottomrule
	 \end{tabular}
	 }%

   \caption{\small{Experimental results (top-5 mAP score) \whfi{on} the TVsum dataset. Our method outperforms all of the compared methods by a large margin.}}
   \label{tab:TVSum}
  
}
\end{table}%

\begin{table}[t] 
	\centering
	   \resizebox{0.9\columnwidth}{!}
		 {
		 \renewcommand\arraystretch{0.9}
	  \begin{tabular}{l|c|c|c|c|c|c|c|c|c|c|c|c|c|c}
			 \toprule
			 \multirow{2}{*}{Topic} &\multicolumn{5}{c|}{Supervised Methods}&\multicolumn{7}{c|}{Weakly supervised} &\multicolumn{2}{c}{Weakly supervised} \\
			 \cline{2-15}
			 &KVS & DPP & sLstm & SM &SMRS & Quasi  & MBF& CVS & SG & LIM-s&VESD& DSN &$\textup{MINI-Net}^\textup{{w/o audio}}$& MINI-Net   \\
			 \hline

			 BJ  &$0.662$&$0.672$&$0.683$&$0.692$&$0.504$&$0.561$&$0.631$&$0.658$&$0.698$&$-$ &$0.685$&$0.715$&$0.7756$& $\mathbf{0.8450}$\\
			 BP  &$0.674$&$0.682$&$0.701$&$0.722$&$0.492$&$0.625$&$0.592$&$0.675$&$0.713$&$-$ &$0.714$&$0.746$&$0.9628$& $\mathbf{0.9887}$ \\
			 ET  &$0.731$&$0.744$&$0.749$&$0.789$&$0.556$&$0.575$&$0.618$&$0.722$&$0.759$&$-$ &$0.783$& $0.813$& $0.7864$&$\mathbf{0.9156}$\\
			 ERC &$0.685$&$0.694$&$0.717$&$0.728$&$0.525$&$0.563$&$0.575$&$0.693$&$0.729$&$-$ &$0.721$& $0.756$&$0.9525$& $\mathbf{1.0000}$\\
			 KP  &$0.701$&$0.705$&$0.714$&$0.745$&$0.521$&$0.557$&$0.594$&$0.707$&$0.729$&$-$ &$0.742$&$0.772$&$0.9585$& $\mathbf{0.9611}$\\
			 MLB &$0.668$&$0.677$&$0.714$&$0.693$&$0.543$&$0.563$&$0.624$&$0.679$&$0.721$&$-$ &$0.687$& $0.727$&$0.8686$& $\mathbf{0.9353}$\\
			 NFL &$0.671$&$0.681$&$0.681$&$0.727$&$0.558$&$0.587$&$0.603$&$0.674$&$0.693$&$-$ &$0.724$&$0.737$& $0.8972$ &  $\mathbf{1.0000}$\\
			 NDC &$0.698$&$0.704$&$0.722$&$0.759$&$0.496$&$0.617$&$0.594$&$0.702$&$0.738$&$-$ &$0.751$& $0.782$&$0.8901$& $\mathbf{0.9536}$\\
			 SL  &$0.713$&$0.722$&$0.721$&$0.766$&$0.525$&$0.551$&$0.624$&$0.715$&$0.743$&$-$ &$0.763$&$0.794$&$0.7865$& $\mathbf{0.8896}$\\
			 SF  &$0.642$&$0.648$&$0.653$&$0.683$&$0.533$&$0.562$&$0.603$&$0.647$&$0.681$&$-$ &$0.674$& $0.709$&$0.7272$& $\mathbf{0.7897}$ \\
			 \hline
			 Average 	&$0.684$&$0.692$&$0.705$&$0.735$&$0.525$&$0.576$&$0.602$&$0.687$&$0.720$&$-$&$0.721$&$0.755$& $0.8605$ & $\mathbf{0.9278}$ \\
			 \bottomrule
			 \end{tabular}
	   }%

	   \caption{\small{Experimental results (top-5 mAP score) \whfi{on} the CoSum dataset. Our method outperforms all of the compared methods by a large margin. The entries with ``-'' mean per-class results are not available for that method.}}
	   \label{tab:CoSum}

 \end{table}%

\begin{figure}[t]
    \begin{center}
        \includegraphics[width=0.9 \linewidth]{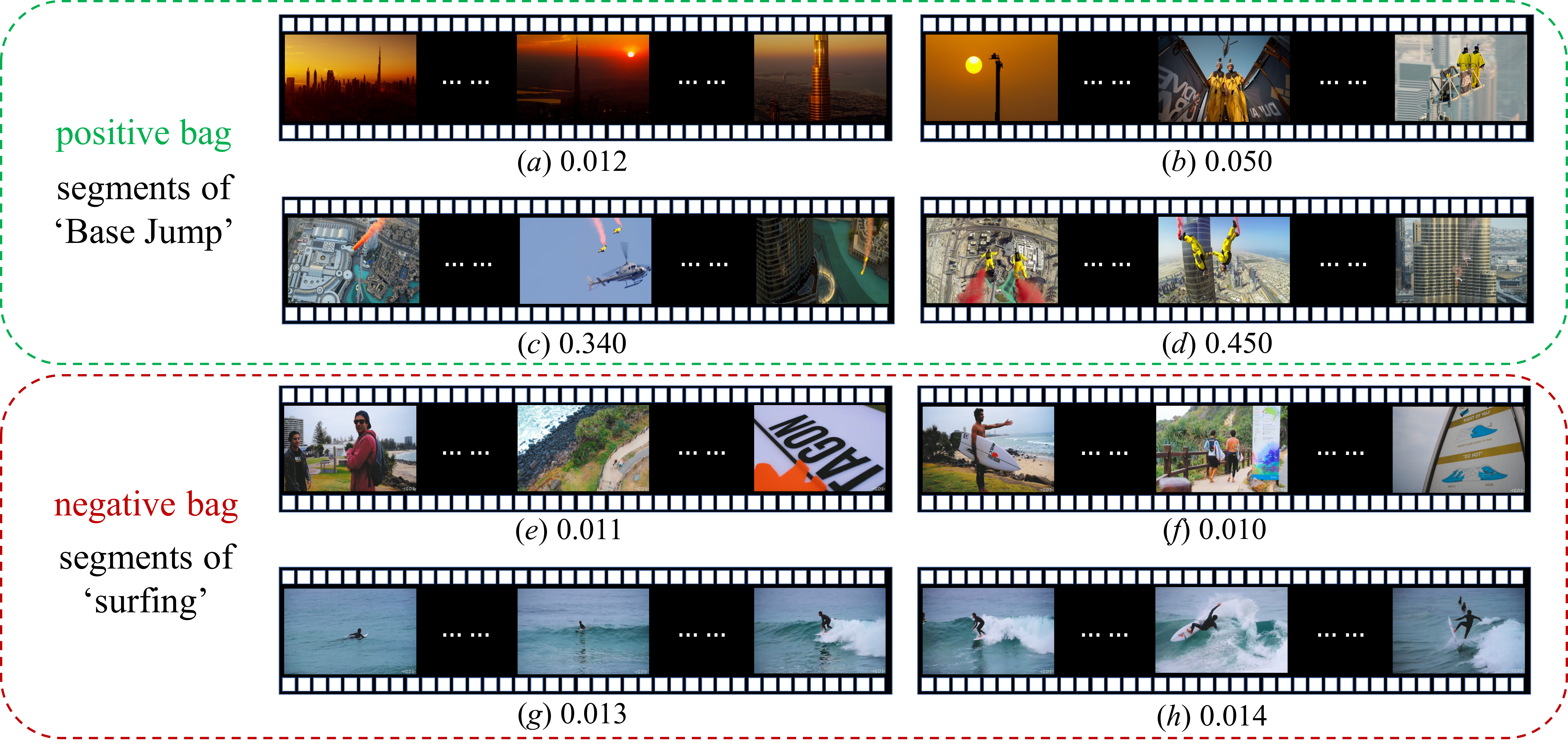}

      \centering\caption{\small{The example of bag in our approach, and the highlight scores \whfi{of} each instance \whfi{estimated} by our MINI-Net trained \whfi{for detecting ``base jump" highlight}. }
      }
   \label{fig:example}
   \end{center}

\end{figure}

\noindent\textbf{- Result on TVSum dataset and CoSum dataset}:
The experimental results for our method on the TVSum \xt{dataset} and the CoSum dataset are shown in Table \ref{tab:TVSum} and Table \ref{tab:CoSum}, respectively. TVsum and CoSum are more challenging datasets as they have diverse videos. However, our method outperforms all of the baselines by a large margin on both the TVSum dataset and the CoSum dataset. Note that LIM-s \cite{xiong2019less}, which is the most competitive ranking-based weakly supervised method, provides the \ftsec{average} top-5 mAP, which is $16.94\%$ less than the value achieved with our MINI-Net on the TVSum dataset. Our approach achieves a significant and consistent improvement over all the events in the two datasets. (\eg, the top-5 mAP of our MINI-Net vs. \xt{that} of VESD are $84.50\%$ vs. $68.5\%$ on the BJ event of CoSum dataset). These results show that the training model based on multiple instance learning using both interest events video data and non-interest events video data is more useful for video highlight detection. As these two datasets consist of long-form videos crawled from social media platforms, in addition to the highlights of the interest event, these videos inevitably contain video information of other events. Figure \ref{fig:example} shows segments and their highlight scores. We can determine that the segments in the non-interest event (\ie, negative bag) are assigned low highlight scores (the segments (e)-(h) in Figure \ref{fig:example}) and the highlights of the interest event achieve the highest scores (the segment (d) in Figure \ref{fig:example}). The performances on the TVsum and CoSum datasets indicate that our model
has the capacity to treat \ftth{segments} from non-interest events as non-highlights and only detect highlights from the interest event.

\noindent\textbf{- Comparison with other multiple instance learning methods.} \ftsec{To further prove that our proposed multiple instance learning framework is suitable for video highlight detection, we compare the other two multiple instance learning frameworks, \ie, Gated-Attention \cite{ilse2018attention} and DMIL-AM \cite{sultani2018real}, which are adapted to video highlight detection. 
It is clearly shown in Table \ref{tab:ablaMIL} that our method performs the best. \eg, MINI-Net outperforms Gate-Attention and DMIL-RM by $17.62\%$ and $13.35\%$ on CoSum dataset, respectively. The results in Table \ref{tab:ablaMIL} demonstrate that the architecture of MINI-Net is more suitable for video highlight detection.
}

\begin{table}[t] 
	\centering
	   \resizebox{0.6\columnwidth}{!}
	{
	\renewcommand\arraystretch{0.9}
	  \begin{tabular}{l|c|c|c}
			 \toprule
			 Dataset& Gated-Attention \cite{ilse2018attention}& DMIL-RM \cite{sultani2018real} & MINI-Net\\
			 \hline
			 YouTube    & $0.6289$ & $0.6357$ & $\mathbf{0.6436}$ \\
			 TVSum      & $0.6533$ & $0.6895$ & $\mathbf{0.7324}$ \\
		     Cosum      & $0.7516$ & $0.7943$  &  $\mathbf{0.9278}$ \\
			 \bottomrule
		\end{tabular}
	   }

	   \caption{\small{Comparisions with related multiple instance learning methods.}}
	   \label{tab:ablaMIL}

\end{table}%

\subsection{Ablation Studies}  \label{sec:ablation}

\ft{We present an ablation study to evaluate each component of our model.}

\noindent\textbf{- Effect of bag modeling.} \ft{Firstly, we evaluate the effect of bag classification module on the proposed model by removing the module, \ie, $\textup{MINI-Net}^\textup{w/o BCM}$}. \ft{Comparing the full model and our model without bag classification module, we clearly observe that the bag classification improves the performance (\eg, ``MINI-Net'' improves the performance of ``$ \textup{MINI-Net}^\textup{{w/o BCM}}$'' from $65.58\%$ to $73.24\%$ for TVSum dataset). This implies that our bag classification module is able to help select as many ground-truth highlights from the video as possible, which benefits video highlight detection.}

\noindent\textbf{- Effect of max-max ranking loss (MM-RL).} \ft{Secondly, we evaluate \ftth{the} impact of MM-RL on our approach. \ftth{$\textup{MINI-Net}^\textup{{w/o MM-RL}}$ indicates that we have removed the MM-RL from the Eq. \ref{eq:finalloss}.} From Table \ref{tab:abla}, we also observe that adding max-max ranking loss can consistently boost the performance (\eg, the results of ``MINI-Net'' vs. those of ``$\textup{MINI-Net}^\textup{{w/o MM-RL}}$'' are $92.78\%$ vs. $77.59\%$ for the CoSum dataset). This result indicates that forcing the most likely highlight segment and the hard non-highlight segment to be far apart in terms of highlight score can help the potential ground-truth highlight segment of the interest event obtain a relatively high score 
.}

\noindent\textbf{- Effect of audio features.} \ft{Finally, to verify that audio is beneficial in our work, we conduct an experiment that trains our model without audio features, \ie, $\textup{MINI-Net}^\textup{w/o audio}$ in Table. \ref{tab:abla}, and $\textup{MINI-Net}^\textup{w/o vision}$ indicates that we have removed the vision feature. More specifically, we use the audio or vision features after several layers of fully connected layers (we make the number of parameters consistent) to replace the fused features that are input to the subsequent network.} \ft{In Table \ref{tab:abla}, we can find that our full method outperforms the alternative variants. In particular, comparing  $\textup{MINI-Net}^\textup{w/o audio}$ and $\textup{MINI-Net}^\textup{w/o vision}$ for the three datasets, the $\textup{MINI-Net}^\textup{w/o vision}$ outperforms $\textup{MINI-Net}^\textup{w/o audio}$  by $9.15\%$, $10.07\%$ and \xt{$16.91\%$} for YouTube Highlights dataset, TVSum dataset and CoSum dataset, respectively. These results indicate that: 1) 
\ftsec{Even using only vision features, our method outperforms the compared methods in Table \ref{tab:YouTube}, Table \ref{tab:TVSum} and Table \ref{tab:CoSum}.}
2) Using audio alone can degrade the performance more than using video alone, as audio is sometimes not native, and music or a voiceover is applied by the \ftsec{video owner}. Such audio cannot be utilized to improve the performance and introduce noise; 3) It is also verified that the combination of audio and vision can improve the performance of the model.}


 \begin{table}[t] 
	\centering
	   \resizebox{0.9\columnwidth}{!}
	{
     \renewcommand\arraystretch{0.9}
	  \begin{tabular}{l|c|c|c|c|c}
			 \toprule
			 Dataset& $ \small{\textup{MINI-Net}}^\textup{{w/o vision}}$& $\small{\textup{MINI-Net}}^\textup{{w/o audio}}$&$ \small{\textup{MINI-Net}}^\textup{{w/o MM-RL}}$&$ \small{\textup{MINI-Net}}^\textup{{w/o BCM}}$&\small{MINI-Net}\\
			 \hline
			 YouTube    & $0.5223$ & $0.6138$ & $0.6166$ & $0.6113$ & $\mathbf{0.6436}$ \\
			 TVSum      & $0.5972$ & $0.6979$ & $0.6495$ & $0.6558$ & $\mathbf{0.7324}$ \\
		     Cosum      & $0.6914$ & $0.8605$ & $0.7759$ & $0.7823$ &  $\mathbf{0.9278}$ \\
			 \bottomrule
\end{tabular}
	   }%
	  
	   \caption{\small{Ablation study on three datasets.}}
	   \label{tab:abla}

 \end{table}%

\section{Conclusion}

 \ft{
  Compared to related work, to our best knowledge, this work is the first to cast the weakly supervised video highlight detection problem as a multiple instance ranking approach. The bag modeling in our multiple instance ranking network (MINI-Net) particularly solves the difficulty of localization of highlight segments of a specific event during training, because MINI-Net works on bag level, where it is only required to ensure a positive bag having a highlight segment of that event and a negative bag having relevant ones. Based on such bag setting, \ws{with a max-max ranking loss}, \ws{our MINI-Net} is able to effectively leverage and quantify all segment information of a video, and therefore the proposed MINI-Net manages to \ws{acquire reliable} higher highlight scores for positive bags as compared to negative bags. The experimental results have validated the effectiveness of our approach.
  }

\section{Acknowledgements}

This work was supported partially by the National Key Research and Development Program of China (2018YFB1004903), NSFC(U1911401,U1811461), Guangdong Province Science and Technology Innovation Leading Talents (2016TX03X157), Guangdong NSF Project (No. 2018B030312002), Guangzhou Research Project (201902010037), and Research Projects of Zhejiang Lab (No. 2019KD0AB03).

\clearpage
%
%
\bibliographystyle{splncs04}
\bibliography{refs}

\newpage
\onecolumn
\section*{\LARGE{Appendix}}\label{s:appendix}
\appendix
\input{sm}

\end{document}

%% file: sm.tex
\section{Implementation Details}

In this work, we implement the proposed method in Pytorch, using SGD as optimizer. The learning rate is initialized as $0.005$ and scaled by a factor of $0.7$ every 20 epochs.
Additionally, we set the weight decay and momentum \whsec{as} $0.0005$ and $0.9$, respectively, for all experiments.
During the training, to be more likely to sample the positive instance,
we take the videos that are shorter than $\tau$ in the interest event as \emph{positive videos} and videos that are longer than $\tau$ in non-interest events as \emph{negative videos}, inspired by \cite{xiong2019less}, \whsec{and we set $\tau$ as $60$. \whsec{W}e set $\epsilon$ as $1$ and the bag size as $60$. We show that our method is not too much sensitive to $\tau$ and the bag size by reporting results of using various value for $\tau$ and the bag size in supplemental experiments.} 
\ftth{\whsec{To form each bag}, we simply break \whsec{a video up uniformly into 1-second segments and randomly sample a bag size number of segments}. If the total number of segments in the video is less than the bag size, we repeat the sampling.}
We follow the standard evaluation metric in \cite{xiong2019less}, \ie, the mean average precision is reported to measure the performance of all of the methods on YouTube Highlights dataset, and top-5 mean average precision for TVSum dataset and CoSum dataset. We adopt \wh{the} C3D network \cite{hara2018can} pretrained on Kinetics \cite{carreira2017quo} to extract a $512$-dimensional feature as vision feature for each segment, \ft{and the VGGish model \cite{hershey2017cnn} pretrained on AudioSet \cite{gemmeke2017audio}} for extracting a $128$-dimensional feature as audio feature.

\section{More Detail about Datasets}
\noindent\textbf{- TVSum} consists of 50 videos grouped by 10 categories (5 videos per category), including changing a Vehicle Tire (VT), getting a Vehicle Unstuck (VU), Grooming an Animal (GA), \xt{Making a Sandwich} (MS), ParKour (PK), PaRade (PR), Flash Mob gathering (FM), BeeKeeping (BK), attempting a \xt{Bike Trick} (BT) and Dog Show (DS).

\noindent\textbf{- CoSum} The dataset consists of 50 videos grouped by 10 categories (5 videos per category), including Base Jumping (BJ), Bike Polo (BP), Eiffel Tower (ET), Excavators River Cross (ERC), Kids Playing in leaves (KP), Major League Baseball (MLB), National Football League (NFL), Notre Dame Cathedral (NDC), Statue of Liberty (SL) and SurFing (SF)

\section{Additional Experimental Results}

\subsection{Variants of max-max ranking loss}
In this work, we exploit a max-max ranking loss (MM-RL) to acquire a reliable relative comparison between the most likely positive segment instance and the most hard negative segment instance. To verify the effectiveness of our proposed MM-RL, we evaluate several variants of our MM-RL.

\noindent\textbf{- Min-Min Ranking Loss}. This variant picks the minimum value from the highlight scores of all segments in both positive bag and negative bag, \ie,  $\min_{\mathcal{I}_{p}^{i} \in \mathcal{B}_p} \mathcal{E}^i_p$ and $\min_{\mathcal{I}_{n}^{i} \in \mathcal{B}_n} \mathcal{E}^i_n$. After that, the min-min ranking loss ensures that $\min_{\mathcal{I}_{p}^{i} \in \mathcal{B}_p} \mathcal{E}^i_p$ is larger than $\min_{\mathcal{I}_{n}^{i} \in \mathcal{B}_n} \mathcal{E}^i_n$ with a margin of $\epsilon$ as follows:
 \begin{equation}\label{eq:min-min}
\begin{aligned}
    \mathcal{L}_{min-min}(\mathcal{B}_p,\mathcal{B}_n) = \max(0,\epsilon-\min_{\mathcal{I}_{p}^{i} \in \mathcal{B}_p} \mathcal{E}^i_p + \min_{\mathcal{I}_{n}^{i} \in \mathcal{B}_n} \mathcal{E}^i_n)
\end{aligned}
\end{equation}

\noindent\textbf{- Min-Max Ranking Loss}. Differently, 
min-max ranking loss, a variant of our max-max ranking loss, picks the minimum value and maximum from the highlight scores of all segments in the positive bag and negative bag, respectively (\ie,  $\min_{\mathcal{I}_{p}^{i} \in \mathcal{B}_p} \mathcal{E}^i_p$ and $\max_{\mathcal{I}_{n}^{i} \in \mathcal{B}_n} \mathcal{E}^i_n$).
After that, Min-min ranking loss ensures that $\min_{\mathcal{I}_{p}^{i} \in \mathcal{B}_p} \mathcal{E}^i_p$ is larger than $\max_{\mathcal{I}_{n}^{i} \in \mathcal{B}_n} \mathcal{E}^i_n$ with a margin of $\epsilon$ as follows:
 \begin{equation}\label{eq:min-max}
\begin{aligned}
    \mathcal{L}_{min-max}(\mathcal{B}_p,\mathcal{B}_n) = \max(0,\epsilon-\min_{\mathcal{I}_{p}^{i} \in \mathcal{B}_p} \mathcal{E}^i_p + \max_{\mathcal{I}_{n}^{i} \in \mathcal{B}_n} \mathcal{E}^i_n)
\end{aligned}
\end{equation}

\noindent\textbf{- Max-Min Ranking Loss}. Moreover, we also evaluate the max-min ranking loss variant that picks the maximum value and minimum from the highlight scores of all segments in the positive bag and negative bag, respectively (\ie,  $\max_{\mathcal{I}_{p}^{i} \in \mathcal{B}_p} \mathcal{E}^i_p$ and $\min_{\mathcal{I}_{n}^{i} \in \mathcal{B}_n} \mathcal{E}^i_n$) before ensuring that $\max_{\mathcal{I}_{p}^{i} \in \mathcal{B}_p} \mathcal{E}^i_p$ is larger than $\min_{\mathcal{I}_{n}^{i} \in \mathcal{B}_n} \mathcal{E}^i_n$ with a margin of $\epsilon$:

 \begin{equation}\label{eq:max-min}
\begin{aligned}
    \mathcal{L}_{max-min}(\mathcal{B}_p,\mathcal{B}_n) = \max(0,\epsilon-\max_{\mathcal{I}_{p}^{i} \in \mathcal{B}_p} \mathcal{E}^i_p + \min_{\mathcal{I}_{n}^{i} \in \mathcal{B}_n} \mathcal{E}^i_n)
\end{aligned}
\end{equation}
 \begin{table}[ht] 
	\centering
	   \resizebox{1\columnwidth}{!}
	{
	  \begin{tabular}{l|c|c|c|c}
			 \toprule
			 Dataset& $ \textup{MINI-Net}^\textup{{Min-Min}}$& $\textup{MINI-Net}^\textup{{Min-Max}}$&$ \textup{MINI-Net}^\textup{{Max-Min}}$& MINI-Net\\
			 \hline
			 YouTube    & $0.5884$ & $0.6165$ & $0.6186$ & $0.6436$\\
			 TVSum      & $0.6469$ & $0.6747$ & $0.7103$ & $0.7324$  \\
		     Co\xt{S}um      & $0.7863$ & $0.8004$ & $0.8338$ & $0.9278$ \\
			 \bottomrule
        \end{tabular}
	   }%
	   \vspace{0.1cm}
	   \caption{Ablation study for ranking loss on three datasets. }
	   \label{tab:ablaRL}
	\vspace{-0.5cm}
 \end{table}%

We adopt three variants mentioned above into our MINI-Net by replacing max-max ranking loss (\ie, $ \textup{MINI-Net}^\textup{{Min-Min}}$ for $\mathcal{L}_{min-min}$, $ \textup{MINI-Net}^\textup{{Min-Max}}$ for $\mathcal{L}_{min-max}$ and $ \textup{MINI-Net}^\textup{{Max-Min}}$ for $\mathcal{L}_{max-min}$). We evaluate these variants for highlight detection on three datasets (\ie, YouTube Highlights dataset, TVSum dataset and CoSum dataset.) and report our experimental results on Table \ref{tab:ablaRL}. From Table \ref{tab:ablaRL}, our proposed MINI-Net with max-max ranking loss performs the best, followed by $ \textup{MINI-Net}^\textup{{Max-Min}}$, for the reason that picking $\max_{\mathcal{I}_{p}^{i} \in \mathcal{B}_p} \mathcal{E}^i_p$ can ensure that the highlight segment of interest event is selected with the highest probability, and picking $\max_{\mathcal{I}_{n}^{i} \in \mathcal{B}_n} \mathcal{E}^i_n$ means that all segments from non-interest events is non-highlights.

\subsection{Evaluation of hyperparameters}

\begin{figure}[ht]
    \begin{center}
        \includegraphics[width=1 \linewidth]{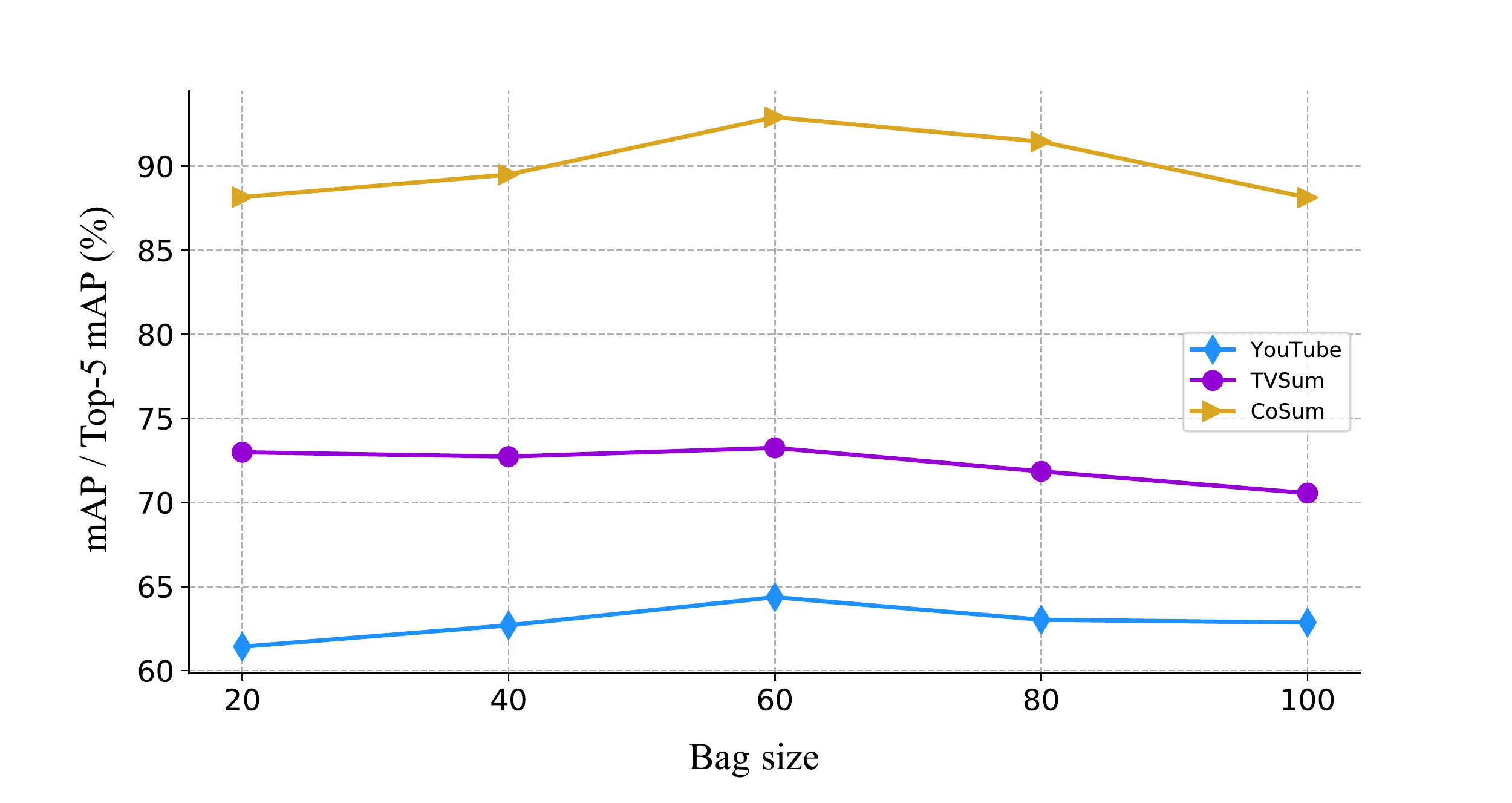}
        \vspace{-0.7cm}
      \centering\small\caption{Accuracy vs. bag size of our multiple instance learning framework on three datasets.
      }
    \label{fig:bagsize}
    \vspace{-0.5cm}
    \end{center}
\end{figure}

\ftsec{Figure \ref{fig:bagsize} shows highlight detection accuracy as a function of bag size. We conduct this ablation on three datasets, \ie, YouTube Highlights dataset, TVSum dataset and CoSum dataset. It can be seen that our method has little performance variance on the three datasets as increasing the number of bag size.}

 \begin{table}[ht] 
	\centering
	   \resizebox{0.5\columnwidth}{!}
	{
	  \begin{tabular}{l|c|c|c}
			 \toprule
			 Dataset& $ \tau = 40$& $\tau = 60$&$ \tau = 80$ \\
			 \hline
			 YouTube    & $0.6150$ & $0.6436$ & $0.6290$ \\
			 TVSum      & $0.7003$ & $0.7324$ & $0.6981$ \\ 
		     Cosum      & $0.8622$ & $0.9278$ & $0.8825$ \\
			 \bottomrule
        \end{tabular}
	   }%
	   \vspace{0.1cm}
	   \caption{Evaluation of different $\tau$ \xt{set} in training process on three datasets. }
	   \label{tab:tau}
	\vspace{-0.5cm}
 \end{table}%
 
 \begin{table}[ht] 
	\centering
	   \resizebox{0.5\columnwidth}{!}
	{
	  \begin{tabular}{l|c|c|c}
			 \toprule
			 Dataset&$ \epsilon = 0.5$& $ \epsilon = 1$&$ \epsilon = 1.5$\\
			 \hline
			 YouTube   & $0.6345$& $0.6436$& $0.6372$\\
			 TVSum     & $0.7174$& $0.7324$& $0.7291$\\ 
		     Cosum     & $0.8999$& $0.9278$& $0.9127$\\
			 \bottomrule
        \end{tabular}
	   }%
	   \vspace{0.1cm}
	   \caption{Evaluation of different $\epsilon$ \xt{set} in training process on three datasets. }
	   \label{tab:epsilon}
	\vspace{-0.5cm}
 \end{table}%
In this work, we take the videos that are shorter than $\tau$ in the interest event as positive videos and videos that are longer than $\tau$ in non-interest events as negative videos. We also conduct the experiments to evaluate threshold $\tau$. Here, we \xt{report} the experimental results on Table \ref{tab:tau}. \xt{It can be found} that our method is not too much sensitive to $\tau$ (\eg, we obtain $1.16\%$ among implementation of $\tau$  = $40$, $60$, $80$ on YouTube Highlights dataset) and we get the best performance when $\tau = 60$. 

We enforce the maximum value of highlight score in positive bag larger than that in negative bag with a margin of $\epsilon$. We test varying value of $\epsilon$, \ie, $0.5$, $1$, $1.5$ and report their results in Table \ref{tab:epsilon}, \xt{it can be found} that our method is not too much sensitive to both $\epsilon$ (\eg, we obtain $0.39\%$ among implementation of $\epsilon$ = 0.5, 1, 1.5 on YouTube Highlights dataset).

\section{Visual Examples}
\begin{figure}[ht]
    \begin{center}
        \includegraphics[width=1 \linewidth]{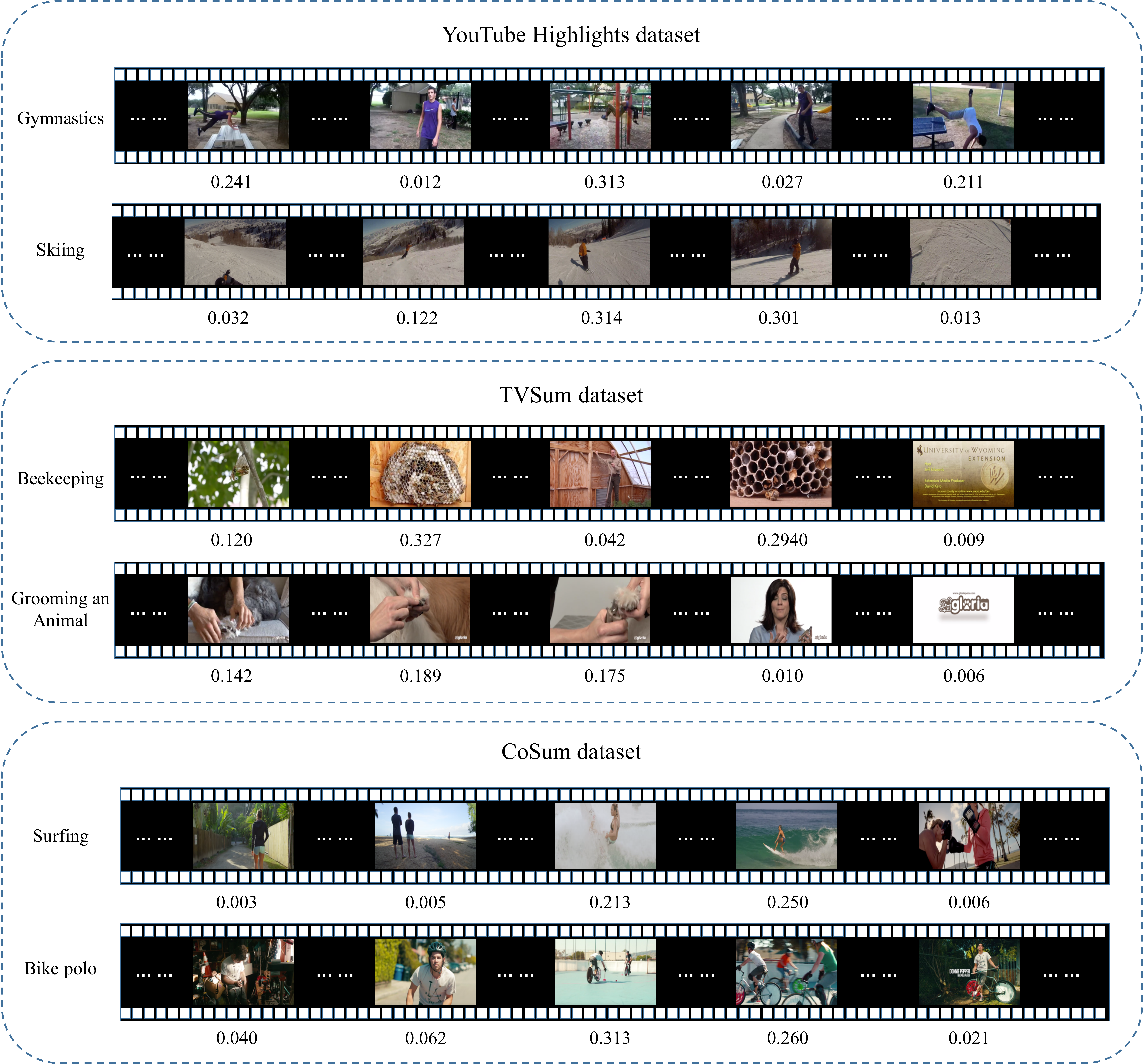}
        \vspace{-0.7cm}
      \centering\small\caption{Examples \xt{of} highlight detection results for three datasets.
      }
    \label{fig:visual}
    \vspace{-0.5cm}
    \end{center}
\end{figure}

Moreover, we also illustrate the highlight detection results on three datasets (\ie, YoutTube Highlights dataset, TVSum dataset and CoSum dataset) in Figure \ref{fig:visual}

\section{Future Discussion}
While we focus on the even-specific highlight detection in this work, our method could be extended to various topics where only weak supervision is provided, including event-agnostic highlight detection. One straightforward idea is \xt{treating} videos which are annotated as highlight-worthy as positive bags and videos with non-highlight-worthy as negative bags for training.